\documentclass{article}

\usepackage{PRIMEarxiv}

\usepackage[utf8]{inputenc} 
\usepackage[T1]{fontenc}    
\usepackage{hyperref}       
\usepackage{url}            
\usepackage{booktabs}       
\usepackage{amsmath,amssymb,amsfonts}       
\usepackage{nicefrac}       
\usepackage{microtype}      
\usepackage{lipsum}
\usepackage{fancyhdr}       
\usepackage{graphicx}       

\title{Layer-wise Relevance Propagation for Echo State Networks applied to Earth System Variability
\thanks{This work was supported by the Helmholtz School for Marine Data Science (MarDATA) funded by the Helmholtz Association (Grant HIDSS-0005).
\textit{\textbf{Citation}}: 
Landt-Hayen, M., Kröger, P., Claus, M., and Rath, W.: "Layer-wise Relevance Propagation for Echo State Networks applied to Earth System Variability", In Proceedings of the 3rd International Conference on Machine Learning Techniques (MLTEC 2022), Zurich, Switzerland, vol. 12, no. 20, pp. 115-130 (2022).}}

\author{
    \textbf{Marco Landt-Hayen} \\
    \textit{GEOMAR Helmholtz Centre for Ocean Research} \\
    Kiel, Germany \\
    mlandt-hayen@geomar.de \\
    \and
    \textbf{Peer Kröger} \\
    \textit{Christian-Albrechts-Universität} \\
    Kiel, Germany \\
    pkr@informatik.uni-kiel.de \\
    \and
    \textbf{Martin Claus} \\
    \textit{Christian-Albrechts-Universität} \\
    Kiel, Germany \\
    mclaus@geomar.de
    \and
    \textbf{Willi Rath} \\
    \textit{GEOMAR Helmholtz Centre for Ocean Research} \\
    Kiel, Germany \\
    wrath@geomar.de
    }

\begin{document}
\maketitle

\begin{abstract}
Artificial neural networks (ANNs) are known to be powerful methods for many hard problems (e.g. image classification, speech recognition or time series prediction). However, these models tend to produce black-box results and are often difficult to interpret. Layer-wise relevance propagation (LRP) is a widely used technique to understand how ANN models come to their conclusion and to understand what a model has learned.

Here, we focus on Echo State Networks (ESNs) as a certain type of recurrent neural networks, also known as reservoir computing. ESNs are easy to train and only require a small number of trainable parameters, but are still black-box models. We show how LRP can be applied to ESNs in order to open the black-box. We also show how ESNs can be used not only for time series prediction but also for image classification: Our ESN model serves as a detector for El Ni\~{n}o Southern Oscillation (ENSO) from sea surface temperature anomalies. ENSO is actually a well-known problem and has been extensively discussed before. But here we use this simple problem to demonstrate how LRP can significantly enhance the explainablility of ESNs.
\end{abstract}

\keywords{Reservoir Computing \and Echo State Networks \and Layer-wise Relevance Propagation \and Explainable AI}

\section{Introduction}
Machine learning (ML) provides powerful techniques in the field of artificial intelligence (AI) to discover meaningful relationships in all kinds of data. Within machine learning, artificial neural networks (ANNs) in shallow and deep architectures are found to be promising and very versatile. While these models considerably push the state-of-the-art solutions of many hard problems, they tend to produce black-box results that are difficult to interpret even by ML experts. Consequently, the question of enhancing the explainability of complex models ("explainable AI" or "xAI") has gained a lot of attention in the AI/ML community and stimulated a large amount of fundamental research \cite{Gilpin2018}, \cite{Ribeiro2016}.
  
In its basic form layers of perceptrons \cite{Rosenblatt1958} are stacked on top of each other to create a multilayer perceptron (MLP) \cite{Ramchoun2016}. These models are usually trained using some form of stochastic gradient descent (SGD) \cite{Ruder2016}. The aim is to minimize some objective or loss function. More sophisticated architectures e.g. make use of convolutional neural networks (CNNs) \cite{Shea2015} or long short term memory (LSTM) \cite{Hochreiter1997} units to have recurrence in time in so-called recurrent neural networks (RNNs).

In this paper, we focus on geospatial data, which typically feature non-linear relationships among observations. In this szenario, ANNs are good candidate models, since ANNs are capable of handling complex and non-linear relations by learning from data and training some adjustable weights and biases \cite{Goodfellow2016}. In recent years these methods have been used in various ways on geospatial data \cite{Mahesh2019}, \cite{Kiwelekar2020}, \cite{Kim2022}.

The problem with using ANNs on data of the Earth system is that we often only have relatively short time series to predict on or a small number of events to learn from. Using sophisticated neural networks encounters a large number of trainable parameters and these models are prone to overfitting. This requires a lot of expertise and effort to train these models and prevent them from getting stuck in local minima of the objective function. Famous techniques are dropout, early stopping and regularization \cite{Srivastava2014}, \cite{Bai2021}, \cite{Krogh1991}.

In this work we overcome these problems by using Echo State Networks (ESNs) \cite{Sun2020}. ESNs are a certain type of RNNs and have been widely used for time series forecasting \cite{Kim2020}, \cite{Gallicchio2017}. In its basic form an ESN consists of an input and an output layer. In between we find a reservoir of sparsely connected units. Weights and biases connecting inputs to reservoir units and internal reservoir weights and biases are randomly initialized. The input length determines the number of recurrent time steps inside the reservoir. We record the final reservoir states and only the output weights and bias are trained. But opposed to other types of neural networks, this does not encounter some gradient descent methods but is rather done in a closed-form manner by applying linear regression of final reservoir states onto desired target values to get the output weights and bias.

This makes ESN models extremely powerful since they require only a very small number of trainable parameters (the output weights and bias). In addition to that, training an ESN is easy, fast and leads to stable and reproducible results. This makes them especially suitable for applications in the domain of climate and ocean research.

But as long as ESNs remain black-boxes, there is only a low level of trust in the obtained results and using these kinds of models is likely to be rejected by domain experts. This can be overcome by adopting techniques from computer vision developed for image data to climate data. Layer-wise relevance propagation (LRP) is a technique to trace the final prediction of a multilayered neural network back through its layers until reaching the input space \cite{Bach2015}, \cite{Montavon2017}. When applied to image classification, this reveals valuable insights in which input pixels have the highest relevance for the model to come to its conclusion. 

Toms, Barnes and Ebert-Uphoff have shown in their work \cite{Toms2020} that LRP can be successfully applied to MLP used for classification of events related to some well-known Earth system variablity: El Ni\~{n}o Southern Oscillation (ENSO).

This work is inspired by \cite{Toms2020} and goes beyond their studies: We also pick the well-known ENSO problem \cite{Ropelewski1986}. ENSO is found to have some strong zonal structure: It comes with anomalies in the sea surface temperature (SST) in Tropical Pacific. This phenomenon is limited to a quite narrow range of latitude and some extended region in terms of longitude. We use ESN models for image classification on SST anomaly fields. We then open the black-box and apply LRP to ESN models, which has not been done before - to the best of our knowledge. 

SST anomaly fields used in this work are found to be noisy. For this reason we focus on a special flavour of ESNs, that uses a leaky reservoir because they have been considered to be more powerful on noisy input data, compared to standard ESNs \cite{Jaeger2007a}. With the help of our LRP application to ESNs, we find the leak rate used in reservoir state transition to be a crucial parameter determining the memory of the reservoir. Leak rate needs to be chosen appropriately to enable ESN models to reach the desired high level of accuracy.

Our models yield competitive results compared to linear regression and MLP used as baselines. However, ESN models require significantly less parameters and hence prevent our model from overfitting. We even find our reservoirs to be robust against random permutation of input fields, destroying the zonal structure in the underlying ENSO anomalies. 

This opens the door to use ESNs on unsolved problems from the domain of climate and ocean science and apply further techniques of the toolbox of xAI \cite{Simonyan2014}.

The rest of this work is structured as follows: In Section \ref{section:ESN} we briefly introduce basic ESNs and focus on reservoir state transition for leaky reservoirs. We then sketch an efficient way to use ESN models for image classification. Section \ref{section:LRP} outlines the concept of LRP in general before we customize LRP for our base ESN models by unfolding the reservoir recurrence. The classification of ENSO patterns and the application of LRP to ESN models is presented in Section \ref{section:Application}. Our models are not only found to be competitive classifiers but also reveal valuable insights in what the models have learned. We show robustness of our models on randomly permuted input samples and visualize how the leak rate determines the reservoir memory. Discussion and conclusion is found in Section \ref{section:Discussion}, followed by technical details on the used ESN and baseline models in the Appendix.

\section{Echo State Networks} \label{section:ESN}
An ESN is a special type of RNNs and comes with a strong theoretical background \cite{Sun2020}, \cite{Jaeger2001}, \cite{Jaeger2002}. ESN models have shown outstanding advantages over other types of RNNs that use gradient descent methods for training. We use in this work a shallow ESN architecture consisting of an input and output layer. In between we find a single reservoir of sparsely connected units. The weights connecting input layer and reservoir plus the input bias terms are randomly initialized and kept fixed afterwards. We find some recurrence within the reservoir and reservoir weights and biases are also randomly set and not trainable. Reservoir units are sparsely connected with sparsity usually in the range of 20-30\%. Further constraints are put to the largest Eigenvalue of the reservoir weight matrix $W_{res}$. This is required for the reservoir to be stable and show the so-called Echo State Property \cite{Jaeger2007}.

\begin{figure}[t]
\centering
\includegraphics[width=0.50\textwidth]{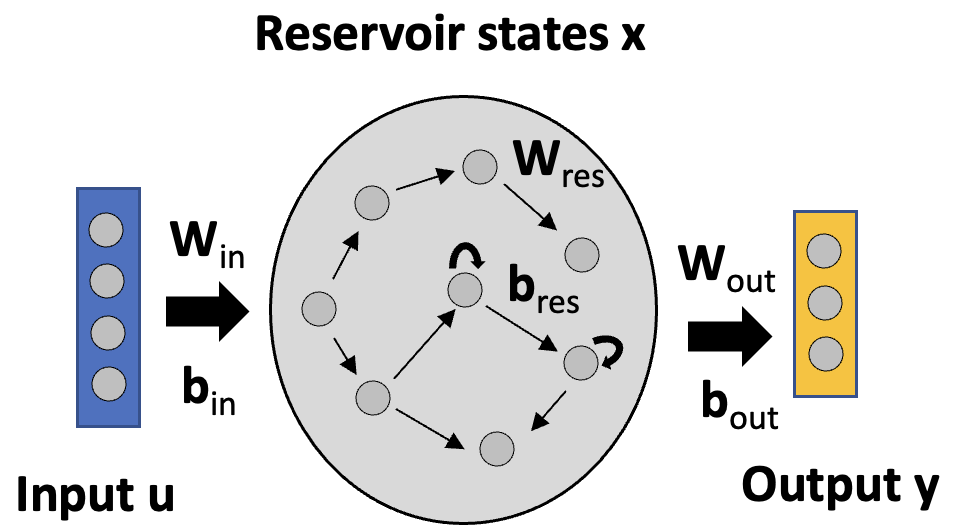}
\caption{Sketch of base ESN: An input and an output layer, in between we find the reservoir.}
\label{fig:baseESN}
\end{figure}

Only the output weights and bias are trained by solving a linear regression problem of final reservoir states onto desired target outputs. A sketch of a base ESN model is shown in Figure \ref{fig:baseESN}. 

In our ESN model, $u(t) \in \mathbb{R}^{D \times 1}$ denotes input values at time $t$ with $D$ input features. Inputs are fed into the model for $T$ time steps, hence $t=1..T$. Reservoir states at time $t=1..T$ are denoted by $x(t) \in \mathbb{R}^{N \times 1}$, final reservoir states are obtained as $x(T)$. The final model output $y(T) \in \mathbb{R}^{M \times 1}$ at time $T$ has $M$ output values.

We then find input weights $W_{in} \in \mathbb{R}^{N \times D}$, connecting $D$ input units to $N$ reservoir units. Reservoir weights are given by $W_{res} \in \mathbb{R}^{N \times N}$ and output weights connecting $N$ reservoir units to $M$ output units read $W_{out} \in \mathbb{R}^{M \times N}$. In addition to weight matrices we have bias vectors $b_{in} \in \mathbb{R}^{N \times 1}$, $b_{res} \in \mathbb{R}^{N \times 1}$ and $b_{out} \in \mathbb{R}^{M \times 1}$ for input, reservoir and output units, respectively.

We use a leaky reservoir with leak rate $\alpha \in [0,1]$, as discussed in \cite{Jaeger2007a}. Leak rate serves as smoothing constant. The larger the leak rate, the faster reservoir states react to new inputs. In other words the leak rate can be understood as the inverse of the memory time scale of the ESN: The larger the leak rate, the faster the reservoir forgets previous time steps' inputs. The reservoir state transition is defined by Equation \ref{equation:reservoir_state_transition}.

\begin{equation}
    x(t) = (1-\alpha)\,x(t-1) + \alpha \, act[W_{in} u(t) + b_{in} + W_{res} x(t-1) + b_{res}]
  \label{equation:reservoir_state_transition}
\end{equation}

Here $act(.)$ is some activation function, e.g. \emph{sigmoid} or \emph{tanh}. From the initial reservoir states $x(t=1)$ we can then obtain further states $x(t)$ for $t=2..T$ by keeping a fraction $(1-\alpha)$ of the previous reservoir state $x(t-1)$. Current time step's input $W_{in} u(t) + b_{in}$ as well as recurrence inside the reservoir $W_{res} x(t-1) + b_{res}$ are added after applying some activation and multiplying with leak rate $\alpha$. Reservoir states $x(t)$ are only defined for $t=1..T$. This requires special treatment of $x(t=1)$ as outlined in Equation \ref{equation:first_reservoir_state}.

\begin{equation}
x(t=1) = \alpha \, act[W_{in} u(t) + b_{in}]
\label{equation:first_reservoir_state}
\end{equation}

The model output $y(T)$ is derived as linear combination of output weights $W_{out}$ and biases $b_{out}$ with final reservoir states $x(T)$, as shown in Equation \ref{equation:model_output}.

\begin{equation}
y(T) = W_{out}\,x(T) + b_{out}
\label{equation:model_output}
\end{equation}

This is a linear problem that can be solved in a closed-form manner with multi-linear regression minimizing mean squared error to obtain trained output weights and biases.

\begin{figure}[h]
\centering
\includegraphics[width=0.45\textwidth]{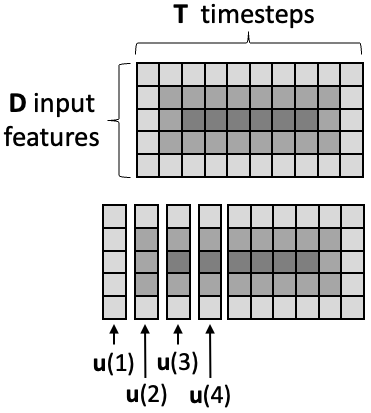}
\caption{In the upper part we show some synthetic 2D input sample consisting of D input features and T time steps. Feeding the sample column by column into the base ESN model requires breaking the sample into columns. In the lower part we show inputs for the first four time steps.}
\label{fig:feedColumnwise}
\end{figure}

ESN models have been widely used for time series forecasting \cite{Kim2020}, \cite{Gallicchio2017}. The idea is to feed a single signal or multiple time series of a specific length $T$ into the model. In our work we want to use 2D image data as input samples. This can be done in various ways. One possibility it to flatten 2D image data to obtain a one-dimensional vector and then couple each input to one reservoir unit in the first time step. Without adding additional inputs the reservoir then swings for some time steps to unfold its dynamics \cite{Woodward2011}. In this approach the number of reservoir units is directly linked to the number of input units. For high dimensional input data reservoirs can hence become quite large. As mentioned above, we need to put some constraint on the largest Eigenvalue of the reservoir weight matrix $W_{res}$ for stability reasons. Getting the largest Eigenvalue becomes computationally intensive for huge reservoirs and we therefore chose a different approach: 

Here we transform images into a temporal signal. This is done by transforming one of the spatial dimensions (longitude) to a temporal one and passing 2D images column-wise into a base ESN model \cite{Schaetti2016}. This is sketched in Figure \ref{fig:feedColumnwise}.

Feeding an image with dimensions $D \times T$ into a base ESN model is equivalent to having $D$ input time series with length $T$. This allows using ESN models for image classification.

\section{Layer-wise Relevance Propagation} \label{section:LRP}
LRP was first introduced by Bach et al. in 2015 \cite{Bach2015}. LRP aims at understanding decisions of non-linear classifiers like ANNs. It can be used on classification and regression problems. This technique opens the black-box by visualizing the contributions of single input units to model predictions. Resulting relevance scores for an individual input sample can be presented as a heat map and give an intuitive understanding of which parts of the input sample have the highest relevance.

LRP has been successfully applied to various network architectures including MLP, CNN or LSTM models \cite{Toms2020}, \cite{Jung2021}. But to the best of our knowledge, LRP has not been used for ESN models.

In this section we will briefly repeat the general idea behind LRP before we customize this technique for using it on base ESN models. 

\begin{figure}[h]
\centering
\includegraphics[width=0.45\textwidth]{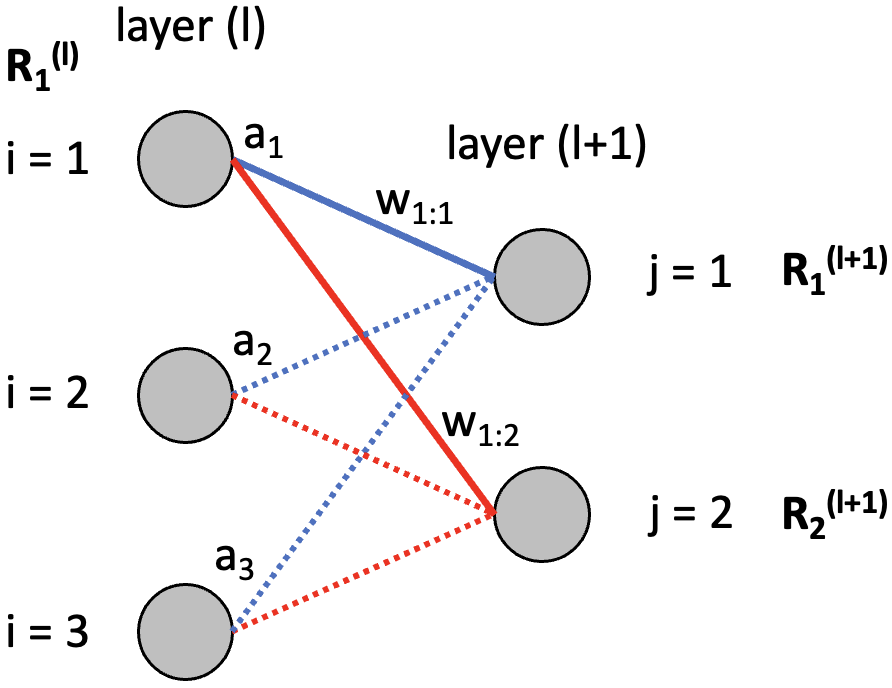}
\caption{Illustrating the general idea behind LRP: Relevance is traced back from higher to lower layers.}
\label{fig:LRP_general}
\end{figure}

\subsection{General idea of LRP} \label{section:LRPgeneral}

LRP, as presented in \cite{Bach2015}, does not provide some closed-form solution but rather comes as a set of constraints. Used on image data it serves as a concept for achieving a pixel-wise decomposition of the final model output $y(T)$, as stated in Equation \ref{equation:LRPpart1}.

\begin{equation}
y(T) = \sum_{n} R_n^{(1)}
\label{equation:LRPpart1}
\end{equation}

The model output $y(T)$ is taken as the \emph{final} or \emph{total} relevance. The ultimate goal is to decompose the final relevance and find the contributions $R_n^{(1)}$, also referred to as \emph{relevance score} of each of the $n$ input pixels. Here superscript $(1)$ refers to the first layer, which is the input layer.

To achieve that goal the relevance is traced back from the output layer all the way through lower layers until we finally reach the input layer. In addition to Equation \ref{equation:LRPpart1} the second constraint is stated in Equation \ref{equation:LRPpart2}.

\begin{equation}
y(T) = \dotsi = \sum_j R_j^{(l+1)} = \sum_i R_i^{(l)} = \dotsi = \sum_{n} R_n^{(1)}
\label{equation:LRPpart2}
\end{equation}

This framework guarantees total relevance to be preserved in each layer. For calculating the relevance map for an individual input sample the trained model weights and biases are fixed. We then start with the model output as final relevance. A common approach for tracing relevance back through lower layers is by taking only positive contributions of pre-activations into account. This clearly satisfies constraints in Equations \ref{equation:LRPpart1} and \ref{equation:LRPpart2}. An example is sketched in Figure \ref{fig:LRP_general}.

Assume units $j=1$ and $j=2$ of layer $(l+1)$ have known relevance scores $R_{1}^{(l+1)}$ and $R_{2}^{(l+1)}$, respectively. This relevance is now distributed on units $i=1, 2, 3$ of layer $(l)$. Unit $i=1$ ends up having relevance $R_{1}^{(l)}$ from two contributions, as stated in Equation \ref{equation:LRPdistribution}: One from unit $j=1$ and one from unit $j=2$ of layer $(l+1)$, indicated by solid blue and red lines in Figure \ref{fig:LRP_general}, respectively. 

\begin{equation}
\begin{aligned}
R_{1}^{(l)} & = \left (\frac{a_1 w_{1:1}}{a_1 w_{1:1} + a_2 w_{2:1} + a_3 w_{3:1}} \right ) R_{1}^{(l+1)}\\
& + \left (\frac{a_1 w_{1:2}}{a_1 w_{1:2} + a_2 w_{2:2} + a_3 w_{3:2}} \right ) R_{2}^{(l+1)} \\
&
\end{aligned}
\label{equation:LRPdistribution}
\end{equation}

Here $a_1$, $a_2$ and $a_3$ denote activations of units $i=1,2,3$ of layer $(l)$, respectively and $w_{i:j}$ denotes the weight connecting some unit $i$ from layer $(l)$ with some unit $j$ from subsequent layer $(l+1)$. 
This can be simplified using $z_{ij}^+ = max(a_iw_{i:j}, 0)$, where $+$ denotes that we only consider positive contributions. Relevance $R_{i=i_0}^{(l)}$ for a unit $i_0$ of layer $(l)$ is stated in Equation \ref{equation:LRPsimplified}.

\begin{equation}
R_{i=i_0}^{(l)} = \sum_j \frac{z_{i_0j}^+}{\sum_i z_{ij}^+} R_{j}^{(l+1)}
\label{equation:LRPsimplified}
\end{equation}

\begin{figure*}[b]
\centering
\includegraphics[width=0.9\textwidth]{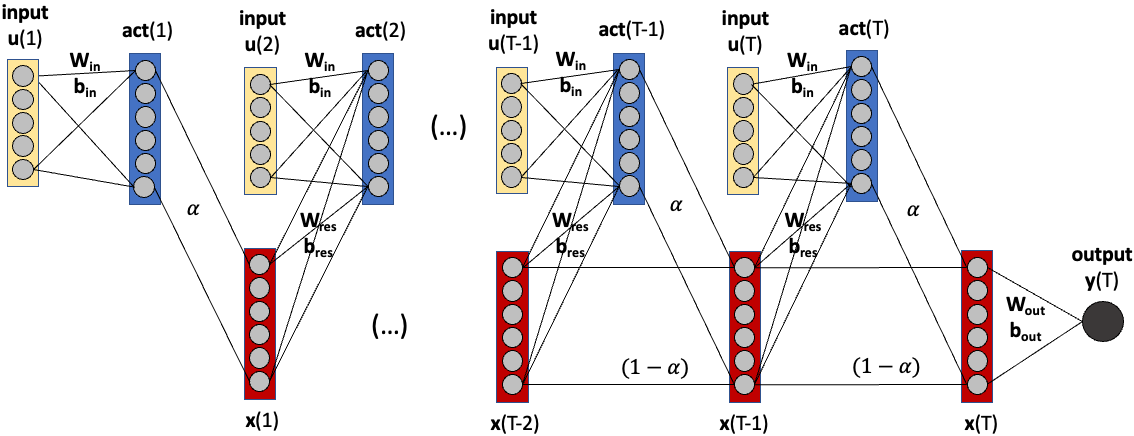}
\caption{Unfolding our base ESN model in time.}
\label{fig:LRP_baseESN}
\end{figure*}

\subsection{LRP customized for ESN models} \label{section:LRPextended}
Applying LRP to ESNs requires extending the basic methodology described in Section \ref{section:LRPgeneral}. Our ESN model consists of an input and an output layer. In between we have the reservoir with recurrence in time. Before we can apply LRP, we need to \emph{unfold} the reservoir dynamics. Feeding an image consisting of $T$ columns into a base ESN model leads to $T$ time steps to be treated as individual layers. Accordingly we have inputs $u(t) \in \mathbb{R}^{D \times 1}$ for time steps $t=1..T$ with $D$ input features.

In Section \ref{section:ESN} we introduced our base ESN model including a \emph{leaky} reservoir with leak rate $\alpha$. Reservoir state transitions for time steps $t=2..T$ have been stated in Equation \ref{equation:reservoir_state_transition}. The initial reservoir states $x(1)$ are somewhat special, since there are no previous time step's reservoir states $x(0)$, as we have seen in Equation \ref{equation:first_reservoir_state}. Figure \ref{fig:LRP_baseESN} shows unfolded reservoir dynamics. In addition to that we decomposed reservoir state transitions to visualize distinct contributions separately. 
As soon as we have trained our base ESN model, the sketch in Figure \ref{fig:LRP_baseESN} can be used to understand how the model output is calculated by forward passing an input sample through the network: In this example we have $D=5$ input features in every time step (shaded yellow), denoted as $u(t)$ for $t=1..T$. For simplicity we further assume to only have six reservoir units providing reservoir states $x(t)$ for $t=1..T$ (shaded red). Reservoir states multiplied with $(1-\alpha)$ contribute to subsequent time step's reservoir states, as sketched in the lower track of Figure \ref{fig:LRP_baseESN}. The second contribution is given by $\alpha\, act(.)$. Here $act(.)$ (shaded blue) is some appropriate activation function (e.g. \emph{sigmoid} or \emph{tanh}) and takes as argument the current time step's input $W_{in} u(t) + b_{in}$ plus incorporates the recurrence inside the reservoir $W_{res} x(t-1) + b_{res}$. Once we calculated final reservoir states $x(T)$ we obtain model output $y(T)$ as seen in Equation \ref{equation:model_output} using trained output weights and bias.

But Figure \ref{fig:LRP_baseESN} also illustrates how LRP works for our base ESN model. As usual, we pick an individual input sample and take the model output as final relevance. We then move backwards through all time steps. Opposed to the \emph{general} concept of LRP, total relevance is not constant from time step to time step. Instead a part of the total relevance is attributed to each time step's input $u(t)$ and only the remaining relevance is passed on until we reach the initial input $u(1)$. The initial input is special in a way, that it absorbs all residual relevance. For $D$ input features we have $u(t) = (u_1(t), u_2(t),..,u_D(t))^T \in \mathbb{R}^{D \times 1}$ for each time step $t=1..T$. And accordingly we obtain relevance scores $R^{(t)} = (R_{1}^{(t)}, R_{2}^{(t)},..,R_{D}^{(t)})^T \in \mathbb{R}^{D \times 1}$. These column vectors of relevance scores $R^{(t)}$ need to be combined to get the final relevance map $R \in \mathbb{R}^{D \times T}$, which can be visualized as heat map having the same dimensions as the input samples.

Thus total relevance is still preserved if we customize LRP to ESN models. However, Equations \ref{equation:LRPpart1} and \ref{equation:LRPpart2} need to be modified and can be combined to Equation \ref{equation:LRPpart3}.

\begin{equation}
y(T) = \sum_{t} R^{(t)} = \sum_{t} \sum_{d} R_{d}^{(t)}
\label{equation:LRPpart3}
\end{equation}

The final model output $y(T)$ is taken as total relevance and equals the sum of relevance scores $R_{d}^{(t)}$ with $t=1..T$ and $d=1..D$. But as mentioned above, the initial input $u(1)$ absorbs all residual relevance. The residual relevance itself depends on the amount of relevance, that has already been attributed to all other time steps' inputs $u(2),..,u(T)$. The speed of decay for total relevance, as it is passed through the layers in a descending order, is controlled by leak rate $\alpha$. The role of $\alpha$ as memory parameter has been discussed in Section \ref{section:ESN}. If $\alpha$ is chosen too low, we find an unreasonably high amount of residual relevance to be assigned to the initial input $u(1)$. To overcome this problem, we add a dummy column of ones as initial column to all input samples. This does not affect model performance, since the additional column is identical for \textit{all} samples. The relevance $R^{(1)}$ attributed to the dummy column is meaningless in the overall relevance map $R$ and can be omitted.

\begin{figure*}[b]
\centering
\includegraphics[width=0.9\textwidth]{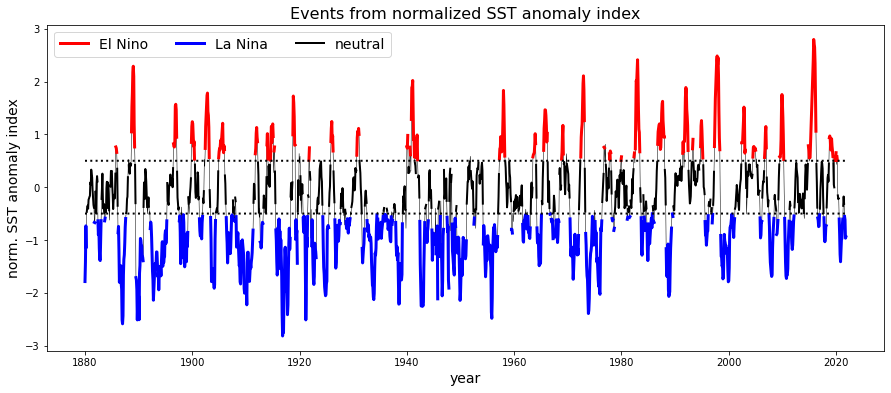}
\caption{SST anomaly index used to define El Ni\~{n}o and La Ni\~{n}a events.}
\label{fig:SSTindex}
\end{figure*}

\section{Application to ENSO} \label{section:Application}
In this section we will briefly recap the main characteristics of ENSO. Additional details on ENSO can be found e.g. in \cite{Toms2020}, \cite{Ropelewski1986}. We then show results from using our base ESN model for classifying 2D input samples and open the black-box by applying LRP as described in Section \ref{section:LRPextended}. We intentionally choose ENSO as well-known problem to gain confidence in our model and methodology to open the door for applying LRP and further xAI techniques with ESN models on unsolved problems in the context of Earth system and climate research. 

\subsection{ENSO patterns} \label{section:ENSOpatterns}
For our studies we use measured monthly mean SST for the years 1880 through 2021, provided by US National Oceanic and Atmospheric Administration. Raw data comes in a 2° by 2° latitude-longitude grid. Each sample consists of $89 \times 180$ grid points.

There are several indices used to monitor the sea surface temperature in the Tropical Pacific. All of these indices are based on SST anomalies averaged across a given region. Usually the anomalies are computed relative to a seasonal cycle estimated from some reference period (climatology) of 30 years (here 1980 through 2009). For our purpose we use SST anomalies averaged over the most commonly used Ni\~{n}o 3.4 region (5°N–5°S, 120–170°W), normalized by its standard deviation over the reference period to obtain a SST anomaly index used to define El Ni\~{n}o and La Ni\~{n}a events, which are associated with anomalous warm and cold SST, respectively. The index is shown in Figure \ref{fig:SSTindex}. El Ni\~{n}o is referred to index values $\ge 0.5$ (red), whereas La Ni\~{n}a events are referred to index values $\le -0.5$ (blue). In between we find \emph{neutral} states, which are not considered here for classification. The SST anomaly index is used for labelling input samples and also as a single continuous target. In the time span from 1880 through 2021 we have a total number of 1,041 samples and split data into train and validation samples, using the first 80\% for training (832 samples) and remaining 20\% for validation (209 samples).

\begin{figure}[h]
\centering
\includegraphics[width=0.45\textwidth]{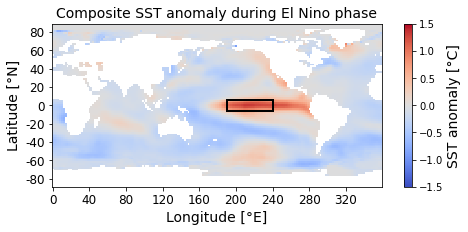}
\includegraphics[width=0.45\textwidth]{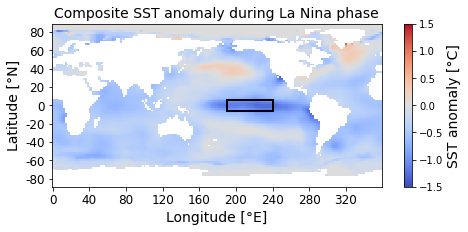}
\caption{Composite average SST anomaly patterns for El Ni\~{n}o (left-hand side) and La Ni\~{n}a (right-hand side) events. Ni\~{n}o 3.4 region is highlighted by a black rectangle.}
\label{fig:ENSOcomposite}
\end{figure}

Composite average SST anomaly patterns for El Ni\~{n}o and La Ni\~{n}a are shown in Figure \ref{fig:ENSOcomposite}.

\subsection{Classification and LRP}
As described in Section \ref{section:ENSOpatterns} we train our models on 832 SST anomaly fields, where each input sample has dimensions $89 \times 180$ (latitude x longitude). SST is not defined over land masses. This reduces the number of \emph{valid} grid points. Raw data shows some unreasonably high or low values: Here we limit SST anomalies to the range of [-5°C, 5°C]. Values exceeding these limits are set to upper and lower bound, respectively.

For our baseline models (linear regression and MLP) we vectorize valid grid points as inputs. SST anomalies are scaled to $[-1,1]$. In any case we use the normalized SST anomaly index shown in Figure \ref{fig:SSTindex} as single continuous target. We then transform this regression problem to fit our classification problem by creating binary predictions from model output: Positive predictions refer to El Ni\~{n}o, whereas negative predictions refer to La El Ni\~{n}a events.

With this setup we easily reach 100\% classification accuracy on both, El Ni\~{n}o and La Ni\~{n}a samples from train and validation data. This perfection was expected, as already shown in \cite{Toms2020} and is due to the simplicity of the underlying problem.

For the base ESN model we do not flatten input samples, as done for the linear regression and MLP approach. Instead we feed 2D SST anomaly fields into our model and use longitude as time dimension. In other words we have 89 input features, each consisting of 180 time steps. We deal with \emph{invalid} grid points by setting SST anomalies to zero after scaling to inputs to $[-1,1]$. Again we use normalized SST anomaly index as single continuous target and create binary predictions from model output. Reservoir's leak rate is set to $\alpha=0.01$.

\begin{figure}[h]
\centering
\includegraphics[width=0.45\textwidth]{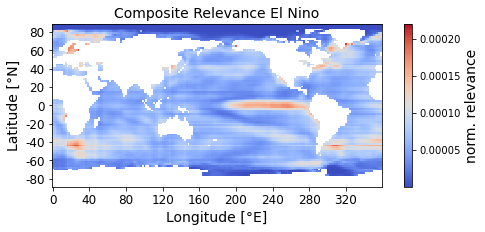}
\caption{Mean relevance (normalized, unitless) obtained from LRP with base ESN model on El Ni\~{n}o train samples.}
\label{fig:meanRelevance}
\end{figure}

This also leads to perfect accuracy on El Ni\~{n}o and La Ni\~{n}a, at least on train data. Validation accuracy is found to be 99\% for both, El Ni\~{n}o and La Ni\~{n}a. We then focus on El Ni\~{n}o, for which we show the mean relevance map obtained from our base ESN model, averaged over \emph{all} train samples in Figure \ref{fig:meanRelevance}.

We find our models to act in a physically reasonable way since highest relevance is put on Ni\~{n}o region in the Tropical Pacific. This appears to be reasonable and shows the model's ability to put its focus on the right spot.

\subsection{Random Permutation}
\begin{figure}[h]
\centering
\includegraphics[width=0.45\textwidth]{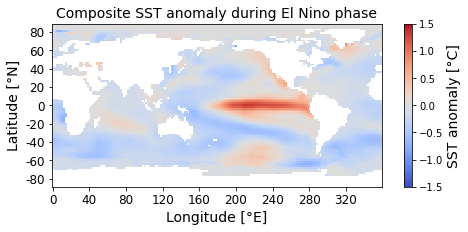}
\includegraphics[width=0.45\textwidth]{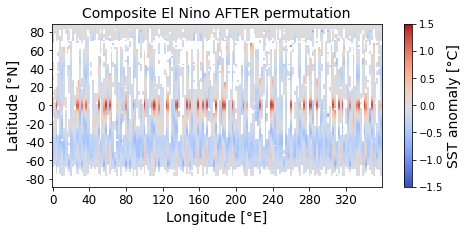}
\caption{Composite average SST anomaly patterns for El Ni\~{n}o (left-hand side) and the same average SST anomaly AFTER some random permutation of columns (right-hand side).}
\label{fig:ElNinoPermutation}
\end{figure}

ENSO patterns show some strong \emph{zonal} structure: SST anomalies for both, El Ni\~{n}o and La Ni\~{n}a, are concentrated on some narrow range in latitude and some extended region in longitude. If we want to use our base ESN model to unknown problems, we need to make sure that this approach is also working without having such characteristic zonal structure present. To proof this, we apply some random (but reversible!) permutation on the columns of \emph{all} input samples before training our model, to \emph{shuffle} the order in time. The result is shown in Figure \ref{fig:ElNinoPermutation}.

\begin{figure}[h]
\centering
\includegraphics[width=0.45\textwidth]{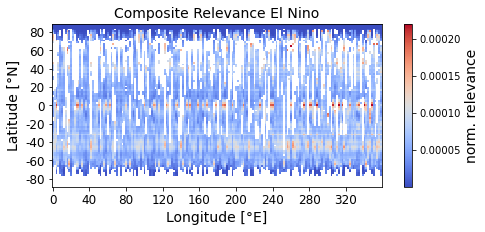}
\includegraphics[width=0.45\textwidth]{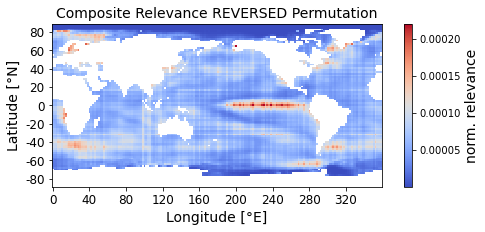}
\caption{Mean relevance (normalized, unitless) obtained from LRP with base ESN model on PERMUTED El Ni\~{n}o train samples (left-hand side). And restored mean relevance after REVERSED permutation (right-hand side).}
\label{fig:LRPpermutation}
\end{figure}

We then train our base ESN model with unchanged parameters and apply LRP on permuted inputs. The obtained mean relevance map calculated on \emph{all} El Ni\~{n}o train samples is shown in Figure \ref{fig:LRPpermutation}. To restore some more familiar mean relevance map, the permutation needs to be reversed. The result is also shown in in Figure \ref{fig:LRPpermutation}.

We find the \emph{restored} mean relevance map to resemble the \emph{original} mean relevance map, shown in Figure \ref{fig:meanRelevance}. This clearly proofs that our approach to pass 2D image data into base ESN models does not rely on the underlying structure in the input data. We also find the same accuracy for base ESN models trained \emph{with} or \emph{without} permuting input columns. This empowers Echo State Networks to be used on unknown problems in the context of climate and ocean science in combination with xAI techniques.

\subsection{Fading Memory}

\begin{figure}[h]
\centering
\includegraphics[width=0.45\textwidth]{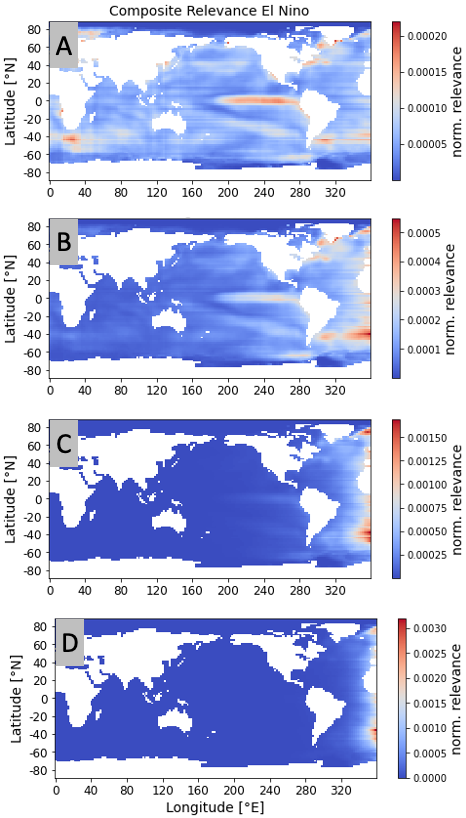}
\caption{Fading memory effect: Mean relevance (normalized, unitless) obtained from LRP with base ESN model on El Ni\~{n}o train samples for $\alpha=0.01$ (A), $0.05$ (B), $0.2$ (C) and $0.4$ (D), respectively.}
\label{fig:FadingMemory}
\end{figure}

In Section \ref{section:ESN} we introduced the reservoir state transition as defined by Equation \ref{equation:reservoir_state_transition}. Leak rate $\alpha$ is found to be a crucial parameter. It determines the memory of the reservoir and can be seen as the inverse of the memory time scale of the ESN: The larger the leak rate, the faster the reservoir forgets previous time steps' inputs. Here we use 2D input samples with $T=180$ time steps for our base ESN model. In other words we feed a 2D input sample column by column into the model, starting on the left-hand side. This procedure requires $\alpha$ to be chosen low enough to enable the reservoir to remember inputs from \emph{all} time steps. This is especially important if we apply our method to unknown problems, since we do not know in advance which time steps are most relevant for achieving optimal performance. 

With increasing leak rate the reservoir memory fades. This effect is visualized in Figure \ref{fig:FadingMemory}. Here we show mean relevance maps for El Ni\~{n}o obtained from ESN models trained with four different leak rates $\alpha=0.01$, $0.05$, $0.2$ and $0.4$, respectively. For $\alpha=0.01$ and $0.05$ we find classification accuracy on train samples to be 100\%, while validation accuracy reaches 99\%. Accordingly we observe high relevance in the Tropical Pacific region, as seen in relevance maps (A) and (B) in Figure \ref{fig:FadingMemory}. This appears to be reasonable for discriminating ENSO patterns. With further increasing $\alpha=0.2$ and $0.4$ the validation accuracy drops to 95\% and 58\%, respectively. Mean relevance maps (C) and (D) in Figure \ref{fig:FadingMemory} explain this decline in model performance: The reservoir simply loses its memory of former input time steps and we find nonzero relevance concentrated on the right-hand side of the relevance maps, representing later time steps. For $\alpha=0.4$ the model fails to distinguish between El Ni\~{n}o and La Ni\~{n}a samples. An accuracy of only 58\% is close to random guessing.

\section{Discussion and Conclusion} \label{section:Discussion}
In this work we successfully used ESNs for image classification and applied LRP to this special type of RNNs, which has not been done before. This enabled us to look inside the model and understand, what the model has learned. LRP is a well-known approach and belongs to the xAI toolbox. Using this technique on a reservoir with $T=180$ time steps is challenging, but possible. Our proposed LRP customized for ESNs also empowers to study the effect of leak rate $\alpha$. We found out that $\alpha$ needs to be chosen appropriately to allow the model to take inputs from \emph{all} time steps into account.

We find accuracy to be competitive compared to baseline models (linear regression and MLP). The advantage of ESNs is the low number of trainable parameters, which makes them fast and, thus, easy to train. In addition, our permutation experiments show, that ESN models yield reproducible and stable results. This even holds true if we only have limited train data, as often in the domain of Earth system and climate research. 

So we can combine the advantages of ESN models with the power of the broad xAI toolbox. Further techniques to be applied to ESN models on similar problems may be backward optimization, sensitivity analysis or salience maps \cite{Toms2020}, \cite{Simonyan2014}.

Beyond application to geospatial data, similar ESN models could be used for time series prediction: Instead of feeding 2D images into the model, we may pass a certain number of climate indices with specific input length to an ESN model and LRP could serve as an alternative for the temporal attention mechanism often used in the context of LSTM sequence-to-sequence models \cite{Bahdanau2014}, \cite{Luong2015}. In this way ESN models have good prospects to help understanding known teleconnections in atmospheric science or to find new relationships \cite{Pak2014}, \cite{Park2018}, \cite{Zhang2019}.

\section*{Appendix: Model Details} \label{section:Appendix}
In this section we briefly present some technical details on the multilayer perceptron used as baseline model and on our ESN model. The MLP was trained on vectorized SST anomaly fields, where we only considered valid grid points. In this case we worked with 10,988 input values for each sample. The input layer of the MLP consists of the same number of input units. We then have two hidden layers of 8 units each and finally one output unit. For a fully connected MLP we end up with 87,993 trainable weights and biases. We used a linear activation function (identity) for all layers and the Adam optimizer \cite{Kingma2014} with constant learning rate $lr=0.0005$. The model was trained over 30 epochs with a batch size of 10. Since we have a regression problem using continuous SST anomaly index as single target, we took the mean squared difference of model output and ground truth as loss function, also referred to as mean squared error loss.

For our base ESN model the number of reservoir units is set to $n_{res}=300$. Input and reservoir weights and biases are drawn from a random uniform distribution in $[-0.1,0.1]$. Reservoir units are only sparsely connected with $sparsity=0.3$. After initialization the reservoir weights are normalized: The largest Eigenvalue of the reservoir weight matrix is set to $0.8$. Leak rate is set to $\alpha=0.01$. As activation in the reservoir state transition we use \emph{tanh}. With this setup our base ESN model only requires 300 trainable output weights plus one output bias, which is significantly less compared to 87,993 trainable parameters for the MLP model.

Raw data used in this work has been uploaded to Zenodo \cite{LandtHayen2022}. Annotated Python code can be found in our GitHub repository \cite{LandtHayen2022a}.

\end{document}